\documentclass[a4paper, 10pt, conference]{ieeeconf}      % Use this line for a4 paper

\IEEEoverridecommandlockouts                              % This command is only needed if 
\usepackage{graphics} % for pdf, bitmapped graphics files
\usepackage{epsfig} % for postscript graphics files
\usepackage{mathptmx} % assumes new font selection scheme installed
\usepackage{times} % assumes new font selection scheme installed
\usepackage{amsmath} % assumes amsmath package installed
\usepackage{amssymb}  % assumes amsmath package installed
\usepackage{subcaption}
\begin{document}
\title{\LARGE \bf
Graph Convolution Networks for Probabilistic Modeling of Driving Acceleration
}

\author{Jianyu Su$^{1}$, Peter A. Beling$^{2}$, Rui Guo$^{3}$, and Kyungtae Han$^{4}$% <-this % stops a space
%\thanks{*Jianyu Su completed this work as an intern in Toyota InfoTech Labs}% <-this % stops a space
\thanks{$^{1}$Jianyu Su is a doctoral student in the Department of Engineering Systems and Environment,
        University of Virginia, 151 Engineer's Way, Charlottesville, VA, 22904, U.S.A
        {\tt\small js9wv@virginia.edu}}%
\thanks{$^{2}$Peter A. Beling is a professor in the Department of Engineering Systems and Environment,
        University of Virginia, 151 Engineer's Way, Charlottesville, VA, 22904, U.S.A
        {\tt\small pb3a@virginia.edu}}%
\thanks{$^{3}$Rui Guo is a principal researcher in Toyota InfoTech Labs,
        465 N Bernardo Ave, Mountain View, CA, U.S.A
        {\tt\small rui.guo@toyota.com}}%
\thanks{$^{4}$Kyungtae Han is a principal researcher in Toyota InfoTech Labs,
        465 N Bernardo Ave, Mountain View, CA, U.S.A
        {\tt\small kyungtae.han@toyota.com}}%
%\thanks{$^{5}$Prashant Tiwari is a director in Toyota InfoTech Labs,
%        465 N Bernardo Ave, Mountain View, CA, U.S.A
%        {\tt\small prashant.tiwari@toyota.com}}%
}

\maketitle

%%%%%%%%%%%%%%%%%%%%%%%%%%%%%%%%%%%%%%%%%%%%%%%%%%%%%%%%%%%%%%%%%%%%%%%%%%%%%%%%
\begin{abstract}
The ability to model and predict ego-vehicle's surrounding traffic is crucial for autonomous pilots and intelligent driver-assistance systems. Acceleration prediction is important as one of the major components of traffic prediction. This paper proposes novel approaches to the acceleration prediction problem. By representing spatial relationships between vehicles with a graph model, we build a generalized acceleration prediction framework. This paper studies the effectiveness of proposed Graph Convolution Networks, which operate on graphs predicting the acceleration distribution for vehicles driving on highways. We further investigate prediction improvement through integrating of Recurrent Neural Networks to disentangle the temporal complexity inherent in the traffic data. Results from simulation with comprehensive performance metrics support that our proposed networks outperform state-of-the-art methods in generating realistic trajectories over a prediction horizon.
\end{abstract}

\section{Introduction}
Autonomous pilots or intelligent driving assistants predict the future state of traffic in order to warn human drivers about collision risks. The autonomous system in the ego-vehicle should consider not only the ego-vehicle's interactions with its immediate neighbors, but also hierarchical and chains of interactions that might affect the ego-vehicle's future state.

Many approaches have been proposed to predict the behavior of vehicles, with most methods falling into the broad categories of regression formulations or classification formulations. While formulating the problem of predicting vehicle behaviors as a classification problem makes it easier to train the model and compare its performance, this classification approach fails to provide detailed future traffic information for planning the future trajectory. Regression methods, however, are able to infer the future state of  traffic, such as vehicle position, velocity and acceleration. In the literature, many of the methods for the regression formulation of traffic prediction employ Recurrent Neural Networks (RNNs). RNNs are widely used to study time-series data. In particular, researchers have been successfully applying Long-Short Term Memory (LSTM) network to various applications such as speech generation, machine translation, and speech recognition  \cite{hochreiter1997long}. In this work, we also use an RNN structure as part of our proposed framework.

A principal weakness of existing driving behavior prediction methods is that they use models that require inputs of fixed size and fixed spatial organization, making it difficult to generalize from training sets into practice. In \cite{morton2016analysis}, for instance, the proposed method uses a leader-follower model that focuses only on the interactions between the ego-vehicle and its leading vehicle. More recently, neighbor models that capture more interactions between ego-vehicle and its surrounding vehicles have been proposed  \cite{altche2017lstm,lenz2017deep}. Though these neighbor methods show some success in predicting the ego-vehicle's future acceleration, they only consider a fixed number of neighbor vehicles. In addition, they need to deal with information padding if one of the pre-defined neighbors is absent.

Graph neural networks (GNNs) are a type of neural network designed for the analysis of graphs \cite{zhou2018graph}.  Recently, GNNs have been drawing increasing attention from both academia and industry for the flexibility that the graph data structure provides and for their convincing performance on various tasks in different domains, such as social science \cite{hamilton2017inductive,kipf2016semi}, neural science \cite{fout2017protein}, and knowledge graphs \cite{hamaguchi2017knowledge}. For instance, motivated by a first-order approximation of spectral convolution on a graph, Graph Convolution Networks (GCNs) are a computationally efficient variant of GNNs that have shown success in achieving fast and scalable classification of nodes in a graph \cite{kipf2016semi}. Another class of GNNs is the Graph Attention Network (GAT), which utilizes \textit{self-attention} \cite{bahdanau2014neural} to allow for inductive reasoning among nodes, thereby providing additional interpretability while matching other GNNs on benchmark evaluation.

In this paper, we propose a flexible driving behavior prediction framework that we call the  {\em Traffic Graph Framework}. Combining GCNs and LSTMs, our proposed method is able to capture not only spatial features of various sizes but also temporal features. This framework consists of undirected graphs that represent the interactions between vehicles, a multi-layer graph convolution neural network used to directly encode the graph structure, and a fully-connected or LSTM mixture density network used to predict future acceleration distributions. 

In series of empirical tests, we investigate the the performance of our proposed models relative to baselines, including GAT and other GCN variants.  The test environment for our methods is a simulation designed to mimic real-world traffic. The simulation is built using the NGSIM I-80 dataset, which contains vehicle trajectories of more than 2000 individual drivers \cite{colyar2006us}. In the simulation, ego-vehicles' traffic states are propagated based on models' predictions. Models are evaluated by comprehensive metrics to measure the discrepancy between the generated trajectories and the ground truth. Furthermore, ablation studies were performed to analyze the effectiveness of the proposed GCNs and RNN architectures. Results show that including the proposed GCNs and RNN structure improves model's prediction quality.

Our principal contributions are three-fold:
%\begin{quote}
\begin{itemize}
    \item We propose a graph structure to denote vehicle's spatial relationships in a dynamic traffic environment. Our structure supports modeling at fine time scales and can be scaled to include an arbitrary number of neighbors for the ego vehicle.
    \item We introduce new variants of GCN layer-wise propagation rule in the context of traffic modeling and we propose a new acceleration prediction framework combining GCNs and LSTM. We successfully applied our framework to a simulation built from real-world data. The resulting systems outperform others from the literature on the task of acceleration prediction. 
    \item We demonstrate that GAT models fail to make accurate acceleration predictions. This result is significant because GATs have been successful in other traffic modeling settings, notably the work by Diehl et al. \cite{diehl2019graph}. From an investigation of the attention weights generated by the \textit{self-attention} mechanism, we identify the causes of GAT underperformance on our problem.   
\end{itemize}
%\end{quote}

The rest of this paper is organized as follows: In the Related Work section, we summarize prior arts. In the Methodology section, we introduce our framework and our proposed GCN variants. In the Experiment section, we present the training procedure, baselines, and simulation results for all models. In the Discussion section, we elaborate on our findings about GCNs and LSTM in the experiment and demonstrate why GAT fails to generate realistic trajectories. The final section concludes the study.
    
% \textbf{more paragraphs on neural driving models and GCN, mention intelligent driving assistance(IDA)}

% Multi-layer graph convolution neural network applies multiple convolution filters to capture multiple levels of relationships in the graph. Take a two-layer graph convolution neural network as an example, the first convolution filter aggregates information from $h^{0}_{v}$ central node $v_i$ and its neighbor nodes $v_j\in N(v_i)$ to output node $v_i$'s first layer encoding $h^{1}_{v_i}$. The second filter will in turn encode node $v_i$ by $h^{1}_{v_j}$ which contains information about node $v_j$'s neighbors. Compared to other neural driving models, this approach is able to encode multiple layers of relationships recursively by nature.

\section{Related Work}
The task of modeling driving behavior consists of modeling car-following behavior and lane-changing behavior. In our work, we focus on augmenting the car-following model.

Car-following models capture the interaction between the ego-vehicle and the vehicles directly adjacent on the microscopic level of the traffic. Based on the number of interactions captured, models can be categorized  as being either a single-lane or multiple-lanes.

A single-lane model focuses on the interactions between vehicles in a single lane. This model considers up to two kinds of interactions: namely, the ego-vehicle with its leading vehicle, and the ego-vehicle with its following vehicle. Many traditional fixed-form models fall into this category, including the Gazis-Herman Rothery model \cite{chandler1958traffic}, the collision avoidance model \cite{kometani1959dynamic}, linear models \cite{helly1959simulation}, psycho-physical models \cite{michaels1963perceptual}, and fuzzy logic-based models \cite{kikuchi1992car}.

Some recent general driving models have moved away from making assumptions about drivers. Lefèvre et al. compare the performance of feed-forward mixture density network against traditional baselines \cite{lefevre2014comparison}. Their empirical tests suggest that the proposed method is able to achieve performance comparable to the baselines. Morton et al. study the effectiveness of LSTM in predicting driving behavior on highways. They reveal that the LSTM's ability to remember historic states of the ego-vehicle appears to be the key to achieving the state-of-art performance \cite{morton2016analysis}.

More recently, multiple-lane models that consider more interactions, coupled with neural networks, have been introduced in the literature. Kim et al. propose a framework based on LSTM to predict vehicle's future position over the occupancy grid \cite{kim2017probabilistic}. Altche et al. use LSTM that predicts traffic using as input state information on the ego-vehicle states and up to 9 of its neighbors. The model is trained and evaluated on the NGSIM 101 dataset which has trajectories from more than 6000 individual drivers \cite{altche2017lstm}.

Diehl et al. \cite{diehl2019graph} used GNNs for vehicle coordinates prediction and demonstrated that GAT models outperform other baselines. Note that our method differs from that work in three main ways.
First, we are interested in generating vehicle trajectories. Hence, our models are structured to predict $0.1$-second future acceleration, which can be propagated to vehicle trajectories of any length with velocity, acceleration, and coordinates information. Diehl et al. aim to predict the 5-second-later coordinates of a vehicle, which contains limited information for the construction of realistic vehicle trajectories. Second, our framework allows for including an arbitrary number of neighbors. The models of Diehl et al., by contrast, consider only up to 8 neighbors, which might result in ignoring important information about the state of traffic around ego-vehicles. Third, we believe that information of ego-vehicle's past states affects future actions. Diehl et al. does not consider RNN structure, whereas we include this structure because it acts to memorize the past states of a vehicle. Furthermore, we analyze the performance of GNNs with and without RNNs.

%The idea of utilizing GNNs to forecast traffic has been proposed in the literature, with GATs outperforming existing baselines \cite{diehl2019graph}.

\section{Methodology}
This section describes the construction of traffic graphs and our proposed graph convolution network variants.
\subsection{Traffic Graph and Features} 
\begin{figure*}[!htb]
\centering
\begin{subfigure}{0.23\linewidth}
\centering
\includegraphics[width=1\linewidth]{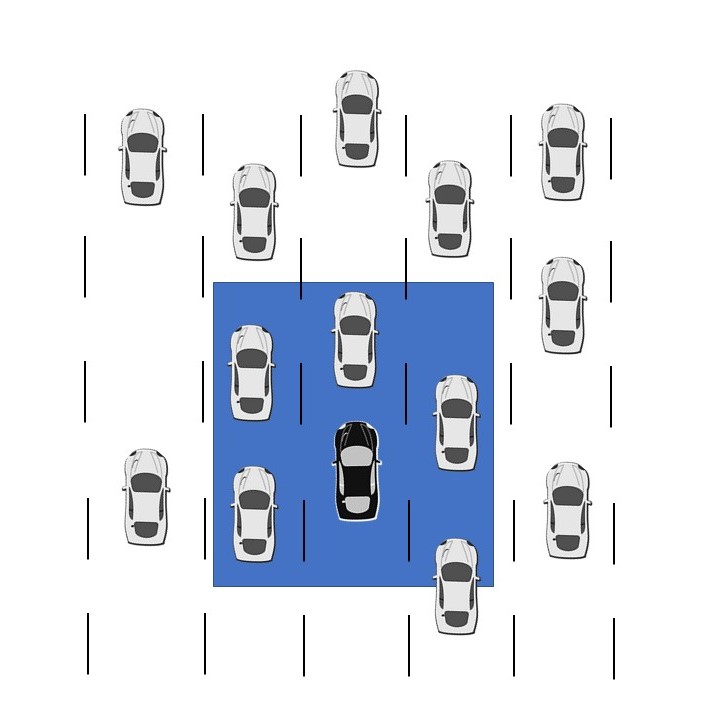} 
\caption{An ego-vehicle considers vehicles only within 1 lanes away as potential neighbor vehicles. A potential neighbor vehicle will be deemed as ego-vehicle's neighbor if and only if the absolute value of their headway distance is smaller than $\tau$}
\label{fig:traffic}
\end{subfigure}
\hfill
\begin{subfigure}{0.23\linewidth}
\centering
\includegraphics[width=1\linewidth, ]{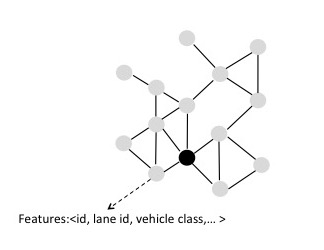}
\caption{A graph is constructed by connecting every vehicle with their neighbor vehicles. Graph nodes share the same feature fields}
\label{fig:graph}
\end{subfigure}
\hfill
\begin{subfigure}{0.23\linewidth}
\centering
\includegraphics[width=1\linewidth]{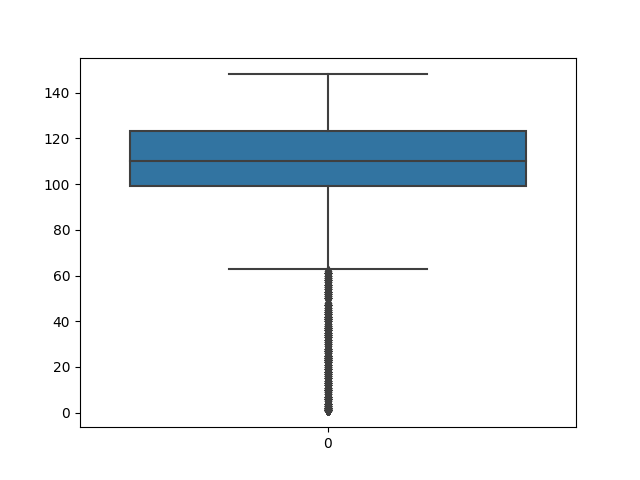} 
\caption{A box plot of the number of nodes in graphs. This depicts the size of traffic graphs}
\label{fig:number}
\end{subfigure}
\hfill
\begin{subfigure}{0.23\linewidth}
\centering
\includegraphics[width=1\linewidth]{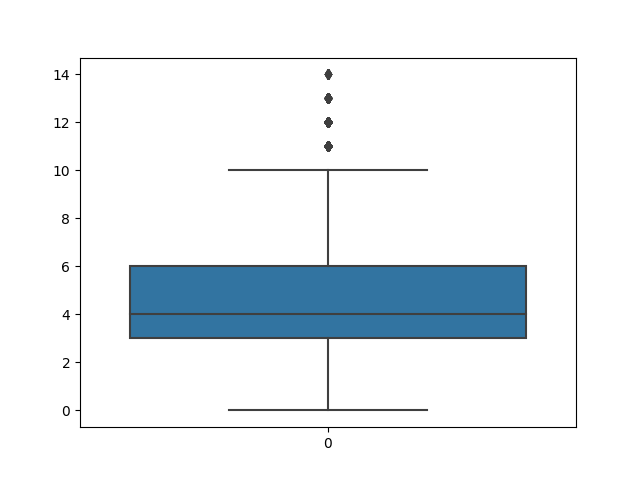} 
\caption{A box plot of the number of neighbours possessed by every vehicle node in graphs. This indicates the number of edges per each node possessed in traffic graphs}
\label{fig:relation}
\end{subfigure}
\caption{Mapping from real world traffic to traffic graph}
\label{fig:traffic2graph}
\end{figure*}
\noindent To leverage the spatial relationships and interactions between vehicles on the highway,  we use an undirected graph $G=(E,V)$ with $N$ nodes $v_i\in V$, edges $(v_i,v_j)\in E$, an adjacency matrix $A \in \mathbb{R}^{N\times N}$, a degree matrix with $D_{ii}=\sum_j A_{ij}$, and a nodes feature information matrix $X \in \mathbb{R}^{N\times F}$ to model the interactions between vehicles. As shown in Figure \ref{fig:traffic2graph}, for a vehicle pair $(v_i,v_j)$ where $v_i\in V$ and $v_j \in V$, the edge $(v_i,v_j)$ is connected if and only if:
%\begin{quote}
\begin{itemize}
    \item vehicle $v_j$ and $v_i$ appear at the same frame; and 
    \item vehicle $v_j$ is less than one lane away from vehicle $v_i$ at the current frame(vehicle $v_j$ should be on the same lane with vehicle $v_i$ or on vehicle $i$'s left, right lanes); and 
    \item the absolute value difference of vehicle $v_j$'s $y$-coordinate and vehicle $v_i$'s $y$-coordinate is less than the designated value $\tau$ at the current frame.
\end{itemize}
%\end{quote}
Note that there is no fixed limit on the number of neighbors; all vehicles within an ego-vehicle's designated distance $\tau$ are its neighbor vehicles. In NGSIM I-80 dataset, the traffic of the study area changes frequently. The traffic, hence, is updated at the same frequency as data was collected in the original dataset. Figure \ref{fig:number}, and Figure \ref{fig:relation} present statistics regarding graphs. 

In this work, we adopt the features used in \cite{lenz2017deep}. For a vehicle node in the graph at frame $t$, its feature vector includes the following elements: vehicle lane id $l_t$, vehicle class id $c$, vehicle velocity $v_t$, vehicle acceleration $a_t$, relative distance from 3 nearest front neighbor vehicles $\{d_{f_1}, d_{f_2}, d_{f_3}\}$ (pad $\tau$ if the number of front neighbors is smaller than 3), and negative relative distance from 3 nearest rear neighbor vehicles $\{-d_{r_1}, -d_{r_2}, -d_{r_3}\}$ (pad $-\tau$ if the number of rear neighbors is less than 3).

\subsection{Graph Convolution Network} 
GCN takes input as a graph $G$ and output nodes encodings. We consider the propagation rule originally introduced in \cite{kipf2016semi} as our base model:
\begin{equation}
\label{eq:orginial gcn}
    H^{l+1}=\sigma \big( \hat{D}^{-\frac{1}{2}}\hat{A}\hat{D}^{-\frac{1}{2}}H^l W^l \big),
\end{equation}
where $\hat{A}=A+I_N$ is the summation of the undirected graph $G$'s adjacency matrix with binary entries $A$ and self-connection $l_N\in \mathbb{R}^{N}$, $l_N\in \mathbb{R}^{N}$ is a identity matrix, $D$ is a degree matrix with $D_{ii}=\sum_j A$, $W^{l}\in \mathbb{R}^{N\times C^l}$ is a matrix of trainable weights at depth $l$, $\sigma$ is an activation function, and $H^{l}$ is the encoding of all nodes in the graph at depth $l$ ($H^0=X$).

This layer-wise propagation rule can be rewritten in the following vector form:
\begin{equation}
    h^{l+1}_{v_i}=\sigma \big(\sum_{j} \frac{h^l_{v_j}}{c_{ij}}W^l +\frac{h^l_{v_i}}{c_{ii}}W^l\big).
\end{equation}
Here, $j$ indexes neighboring nodes of $v_i$, normalization factor $\frac{1}{c_{ij}}$ is an entry located at the $i$th row, $j$th column of $\hat{D}^{-\frac{1}{2}}\hat{A}\hat{D}^{-\frac{1}{2}}$.

The propagation rule represented by Equation \ref{eq:orginial gcn} is a first-order approximation of spectral convolution on a graph. It provides two advantages when used to analyze graphs: first, it enables to aggregate $l^{th}$ order neighborhood of a central node during the encoding process; second, it prevents us from prohibitively expensive eigendecomposition of the graph Laplacian compared with spectral convolution models \cite{kipf2016semi}. Those properties offer us a computational efficient approach to learn the interactions between vehicles that are not directly connected in the graph.

    % \item \textit{LSTM Architecture} We compare models with LSTM architure and their without-LSTM counterparts. Our experiments shows that models with LSTM outperform their counterparts.
\textit{Ego-discriminated GCN} (EGCN): During the implementation of the base model, we find that self-connection affects the performance of the system, an observation that leads to our adaptation of the base model. Self-connection was used to alleviate the problem of vanishing/exploding gradients in GCNs \cite{kipf2016semi}. However, this method applies the same weight $W^l$ to both the central node and its surrounding nodes. In our experiments, we find it is beneficial to remove the self-connection and apply different layer weights to discriminate the central node from its surrounding node. This leaves us with the ego-discriminated propagation rule, which can be represented as follows:

\begin{equation}
    H^{l+1}=\sigma \Big( D^{-\frac{1}{2}}AD^{-\frac{1}{2}}H^lW^l + I_N H^lB^l \Big),
\end{equation}
where $l_N\in \mathbb{R}^{N}$ is an identity matrix, $B^l\in \mathbb{R}^{N\times C^l}$ is trainable weights at depth $l$ for central nodes. The corresponding vector form is given in the following expression:
\begin{equation}
    h^{l+1}_{v_i}=\sigma \Big(\sum_{j} \frac{h^l_{v_j}}{c_{ij}}W^l +\frac{h^l_{v_i}}{c_{ii}}B^l\Big).
\end{equation}

\subsection{Distance-Aware Graph Convolution Network}
For the models mentioned in the previous section, their adjacency matrices $\hat{A}$ and $A$ only denote whether a pair of vehicles is close or not, but they do not describe the degree of closeness. Based on our empirical driving experience--the closer our neighbor vehicle is, the more attention we will pay to it--we use absolute inverse relative distances as entries for our adjacency matrix $\tilde{A}$ to differentiate the degree of closeness between vehicles. Therefore, we introduce the following distance-aware layer-wise propagation rule in our multi-layer GCN (DGCN):

\begin{equation}
\label{eq:dis_matrix}
    H^{l+1}=\sigma \Big( \tilde{D}^{-\frac{1}{2}}\tilde{A}\tilde{D}^{-\frac{1}{2}}H^lW^l+I_N H^lB^l \Big).
\end{equation}
Here, $\tilde{A}$ is an adjacency matrix with $\tilde{A}_{ij}=\frac{1}{|y_{v_i}-y_{v_j}|}$ where $y_i$ represents vehicle $v_i$'s $y$-coordinate. $\tilde{D}$ is a degree matrix with $\tilde{D}_{ii}=\sum_j A_{ij}$. In this propagation rule, $\tilde{A}$'s entries denote the degree of closeness between vehicles. To stablize gradients during training, we discretize the degree of closeness into three levels: $1, 2$, and $3$, which represent far away, medium close and very close, respectively.
Equation \ref{eq:dis_matrix} can also be rewritten in the following vector form:
\begin{equation}
\label{eq:dis_vector}
    h^{l+1}_{v_i}=\sigma \Big(\sum_{j} \frac{h^l_{v_j}}{\tilde{c}_{ij}}W^l +h^l_{v_i}B^l\Big),
\end{equation}
where $\tilde{c}_{ij}$ is an entry located at $i$th row and $j$th column of $\tilde{D}^{-\frac{1}{2}}\tilde{A}\tilde{D}^{-\frac{1}{2}}$. 

\subsection{Gaussian Mixture Model}
In this work, we aim to predict human driver's acceleration distribution given the current traffic state. Hence the output of our network model is Gaussian mixture model (GMM) parameters that characterize the future acceleration distribution. This \textit{mixture density network} (MDN) is first proposed by Bishop \cite{bishop1994mixture} and been successfully applied in speech recognition and other fields \cite{robinson1996use}. For a $K$-component GMM, the probability of the predicted acceleration follows this equation:
\begin{equation}
    p(a)=\sum_{i=1}^{K}w_i \mathcal{N}(a|\mu_i, \sigma_i^2),
\end{equation}
where $w_i, \mu_i$, and $\sigma_i$ are the weight, mean, standard deviation of the $i$th mixture component respectively.
\section{Experiment}
\subsection{Dataset}
The NGSIM I-80 dataset contains detailed vehicle trajectory data collected using synchronized digital video cameras on  eastbound I-80 in Emeryville, CA. This dataset provides precise positions, velocities and other vehicle information over three 15-minute periods at 10 Hz. The study area covers approximately 500 meters in length and consists of six freeway lanes, including a high-occupancy lane and an on-ramp lane. 
We use the NGSIM I-80 reconstructed dataset, which contains vehicles position, velocity, acceleration from 4:00 p.m. to 4:15 p.m., because it corrects errors such as extreme acceleration, and inconsistent vehicle IDs \cite{montanino2013making} \cite{montanino2015trajectory}. We split the data into training sets and testing sets by a ratio of 4 to 1.

\subsection{Data Preparation}
Both training set and testing set are divided into 12-second segments (120 frames). The first 2-second segments (20 frames) are used to initialize the internal state of LSTM networks. Since the aim of the research is to predict driving acceleration using GCNs, we need to prepare traffic graphs from the raw data.

\subsection{Baselines}
Our proposed models are compared with the following non-GNN models and GAT models.

\textit{Fully-connected} (FC): This model shares the same configuration and input features as the GCN without LSTM models.

LSTM: This model's configuration and input features are the same with GCN-LSTM models.

Our proposed models use heuristics to define normalization factors between nodes. For instance, DGCN uses inverse distance as entries for adjacency matrix $A$ based on the heuristic that the ego-vehicles should pay more attention to closer neighbors compared with distant neighbours. In contrast, GAT , which applies \textit{self-attention}, learns to generate the normalization factors for neighbouring nodes rather than resorting to weights in adjacency matrix $A$:
\begin{equation}
    \alpha_{ij} = \frac{\exp{\Big(\text{LeakyReLU}\big(BW_ah_i + WW_ah_j)\big)\Big)}}{\sum_{k \in  N_i}\exp{\Big(\text{LeakyReLU}\big(BW_ah_i + WW_ah_k)\big)\Big)}},
\end{equation}
where $i$ indexes the central node, $j$ indexes the surrounding nodes, $W_a$, $B$ and $W$ are weights in self-attention with $B$ applied to central nodes and $W$ applied to surrounding nodes, LeakyReLU is an activation function, and $\alpha_{ij}$ is equivalent to normalization factor $\frac{1}{c_{ij}}$ mentioned in previous equations. Following the practice in \cite{diehl2019graph}, we utilized different layer weights, $B$ and $W$, to attend to central and neighbouring nodes respectively. \textit{Self-attention} weights $W_a$, $B$, and $W$ are updated such that GAT learns how to distribute $\alpha_{ij}$. 

GAT: This model shares the same configuration and input features as other models without LSTM.

GAT with LSTM: This model's configuration and input features are the same with other LSTM models.
\subsection{Implementation}
All models are trained to output predicted parameters for distributions over future acceleration values. Note that every model in this work shares the same hyperparameters because we aim to compare the effectiveness of GNN and LSTM on improving model performance in the task of driver behavior prediction. We set $\tau=20$ feet, empirically. 

Model structures are shown in Table \ref{table:model configuration}. Each model consists of 3 hidden layers and a 30-component MDN layer. Layer 1 applies Relu activation while other layers do not use any activation. Layer 1 and layer 2 are followed by batch normalization. Batch normalization is a mechanism to address the problem of \textit{internal covariate shift}. It has been reported that adding batch normalization to state-of-the-art image classification networks yields higher classification accuracy compared with the original networks \cite{ioffe2015batch}. The performance of our models is also found to improve when batch normalization is applied. Layer 3 is either an FC layer or an LSTM layer. The final 30-component MDN layer follows layer 3 and has an output size of 90, which corresponds to a 30-component GMM's parameters.

All models are trained for 5 epochs. During training, the models are optimized by the Adam optimizer with a learning rate of $1\times 10^{-3}$ \cite{kingma2014adam}. A dropout of $10$ percent is applied to help prevent overfitting. Gradient norm clipping is also used to deal with gradient vanishing and gradient explosion \cite{pascanu2013difficulty}. All networks are implemented in TensorFlow \cite{abadi2016tensorflow} based on Kpif's \textit{GCN} package \cite{kipf2016semi} and Veli{\v{c}}kovi{\'{c}}'s \textit{GAT} package \cite{velivckovic2017graph}. 
\begin{table*}[h!]
\centering
\caption{Model Configuration}
\begin{tabular}{c | c | c | c | c | c | c | c} 
 \hline
 Model & layer 1 & layer 2 & layer 3 & MDN layer & LSTM & clip norm & adjacency type\\
 \hline
 \hline
 Fully-connected & 128 & 256& 128 & 90 & no& 5 & /\\ 
 GCN base & 128 & 256& 128 & 90 & no& 5 &binary\\
 GAT & 128 & 256& 128 & 90 & no& 5 &binary\\
 EGCN & 128 & 256& 128 & 90 & no& 5 &binary\\
 DGCN& 128 & 256& 128 & 90 & no & 5 &inverse distance\\
 \hline
 LSTM & 128 & 256& 128 & 90 & yes& 5 & /\\
 GCN with LSTM & 128 & 256& 128 & 90 & yes& 5 &binary\\
 GAT with LSTM & 128 & 256& 128 & 90 & yes& 5 &binary\\
 EGCN with LSTM & 128 & 256& 128 & 90 &yes& 5&binary\\
 DGCN with LSTM & 128 & 256& 128 & 90 &yes& 5&inverse distance\\
 \hline
\end{tabular}
\label{table:model configuration}
\end{table*}

\subsection{Evaluation}
Once trained, each model is used to generate simulated trajectories. For every trajectory in the test set, the first 2-second segments (20 frames) of true data are used to initialize LSTM's internal state. In the following 10 seconds, ego-vehicle's velocity and position can be updated by assuming the following equations:
\begin{align}
\label{equation: kinematic}
    v(t+\delta t)=v(t)+a(t+\delta t)\times \delta t\nonumber\\ 
    y(t+\delta t)=y(t)+ v(t+\delta t)\times \delta t,
\end{align}
% \begin{equation}
%     v(t+\delta t)=v(t)+a(t+\delta t)\times \delta\\
%     y(t+\delta t)=y(t)+ v(t+\delta t)\times \delta
% \end{equation}
where $v$ is ego-vehicle's velocity, $y$ is ego-vehicle's $Y$-coordinate and $a$ is vehicle's acceleration.  The graph and node features are updated by propagating other vehicles' true trajectory data and ego-vehicle's simulated trajectory. Following the practice in \cite{morton2016analysis}, we evaluate the quality of simulated trajectories by the following metrics: 
\begin{itemize}
    \item \textit{Root Mean Squared Error (RMSE):} We use root mean squared error to evaluate the discrepancy of speed values between simulated trajectories and true trajectories at designated horizons for a given ego-vehicle:
    \begin{equation}
        RMSE_{velocity}=\sqrt{\frac{1}{mn} \sum_{i=1}^{m}\sum_{j=1}^{n} (v_{H}^i - \hat{v}_{H}^{i,j})^2},
    \end{equation}
    where $m$ is the number of true trajectories, $n=20$ is the number of simulated trajectories per true trajectory, $v_{H}^i$ is the velocity of $i$th true trajectory at horizon $H$, $\hat{v}_{H}^{i,j}$ is the value in $j$th simulated trajectory at time horizon $H$. Similarly, we also use root mean squared error to evaluate the displacement in $Y$-coordinate at 10 second horizon between simulated trajectories and true trajectories:
     \begin{equation}
        RMSE_{Y}=\sqrt{\frac{1}{mn} \sum_{i=1}^{m}\sum_{j=1}^{n} (y^i_{10} - \hat{y}_{10}^{i,j})^2},
    \end{equation}
    where $y_{10}^i$ is the $Y$-coordinate of $i$th true trajectory at 10 second, $\hat{y}_{10}^{i,j}$ is the simulated $Y$-coordinate value for sample $j$ in the $i$th trajectory at 10 second horizon.
    
    \begin{figure}[hbt!]
    \centering
    \includegraphics[width=0.45\textwidth]{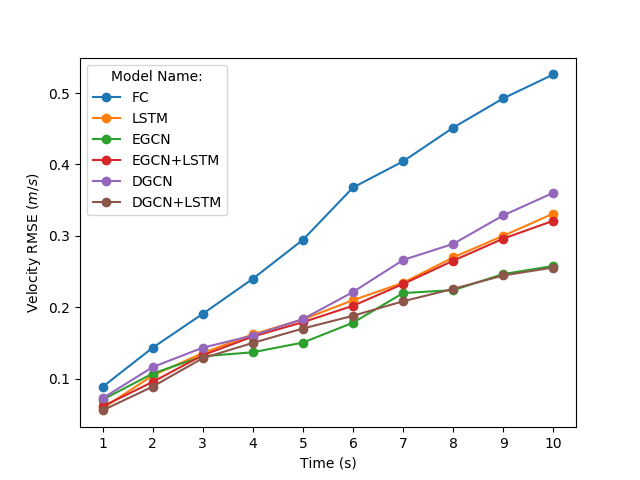}
    \caption{RMSE results for all models}
    \label{fig:overall_mse}
    \end{figure}
    
    Figure \ref{fig:overall_mse} shows the velocity RMSE for the top 6 models over  prediction horizons between 1 and 10 seconds. Models with original GCN \cite{kipf2016semi} and GAT \cite{velivckovic2017graph} are not included because of their bad performance in generating predicted trajectories. In general, the velocity RMSE accumulates over the time horizon. Our adapted GCN models outperform non-GCN models. For non-GCN models, LSTM outperforms the fully-connected model because LSTM is able to access past information. For GCN models, EGCN model and DGCN with LSTM outperform other GCN models. 
    
    The $Y$-coordinate RMSE column in Table \ref{table:rmse} denotes the displacement in $Y$-coordinate between simulated trajectories and their corresponding true trajectories. EGCN model outperforms other models. Velocity RMSE at 10 second horizon reveals the discrepancy of speed between simulated trajectories and the ground truth. DGCN with LSTM outperforms other models in this metric.  
    
    \begin{table*}[!htb]
    \centering
    \caption{RMSE Analysis}
    \begin{tabular}{c | c | c} 
     \hline
     Model & Y RMSE @ 10 s (m)& Velocity RMSE @ 10 s (m/s)\\
     \hline
     
     Fully-connected (FC)&2.89 & 0.526\\ 
     GCN base & 3.52 & 0.622\\
     GAT & 4.13 & 0.688\\
     EGCN & \textbf{1.40} & \textbf{0.258}\\
     DGCN & 1.91 & 0.360\\
     \hline
     LSTM & \textbf{1.61}& 0.331\\
     GCN with LSTM & 3.40&0.653\\
     GAT with LSTM & 4.09 & 0.728\\
     EGCN with LSTM &1.86& 0.321\\
     DGCN with LSTM &1.63&\textbf{0.256}\\
     \hline
    \end{tabular}
    \label{table:rmse}
    \end{table*}
    
    \item \textit{Negative Headway Distance Occurrence:} This metric is used to evaluate models' robustness. It records the occurrences of unrealistic states led by models' poor decision making. Two types of negative headway distances are considered: (1) ego-vehicle's negative headway distance representing collisions with the front vehicle; and (2) following vehicle's negative headway distance denoting collisions between the ego-and its following vehicle. A robust model will have minimal negative headway distance occurrence.
    
    Table \ref{table:jerks} shows the number of negative headway occurrences over number of simulated trajectories for all models. Consistent with RMSE analysis, the results from Table \ref{table:jerks} demonstrates that original GCN models often produce poor acceleration predictions which lead to unrealistic states. EGCN model and DGCN with LSTM model are robust because there are no unrealistic states occurring in their simulated trajectories.
   
    \item \textit{Jerk Sign Inversions:} We use the number of jerk sign inversions per trajectory to evaluate the similarity between the smoothness of the true and simulated trajectories. This metric is used to quantify oscillations in model's acceleration predictions. 
    
    Simulated trajectories of most of models have slightly higher jerk sign inversions than the true trajectories while the LSTM baseline model is not able to generate smooth trajectories. In addition, jerk sign inversions, combined with previous metrics, indicate that the trajectories generated by GAT with LSTM model fail to react against the changes of the ego-vehicle's surrounding traffic.
    
\end{itemize}
\begin{table*}[h!]
\centering
\caption{Jerk Sign Inversions Per Trajectory}
    \begin{tabular}{c | c | c} 
     \hline
     Model & Jerk Sign Inversions & Negative Headway Occurrence Rate\\
     \hline
     
     Fully-connected (FC)& 7.5 & 0.08\\ 
     GCN base & 7.5 & 0.17\\
     GAT & \textbf{5.9} & 0.27\\
     EGCN & 7.5 & \textbf{0}\\
     DGCN & 7.3 & 0.03\\
     \hline
     LSTM & 13.7& 0.02\\
     GCN with LSTM & \textbf{6.7}&0.17\\
     GAT with LSTM & 0.0 & 0.27\\
     EGCN with LSTM &9.5& 0.01\\
     DGCN with LSTM &7.3&\textbf{0}\\
     \hline
     True trajectory& 6.3 & /\\
     \hline
    \end{tabular}
    \label{table:jerks}
\end{table*}

\begin{figure*}[!htb]
\centering
\begin{subfigure}{0.4\linewidth}
\centering
\includegraphics[width=1\linewidth]{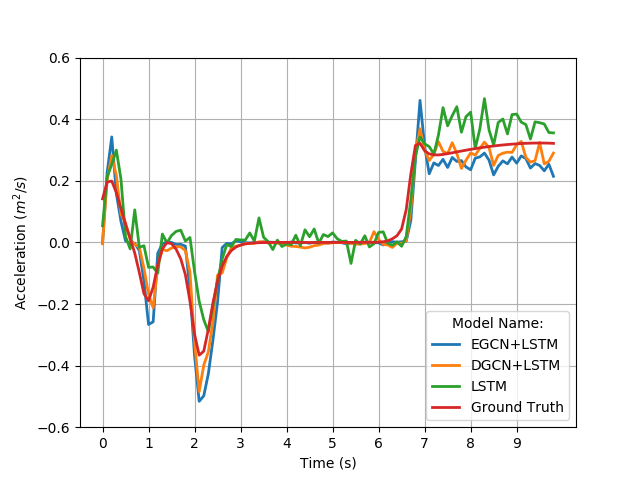} 
\label{fig:subim1}
\caption{LSTM models}
\end{subfigure}
\begin{subfigure}{0.4\linewidth}
\centering
\includegraphics[width=1\linewidth, ]{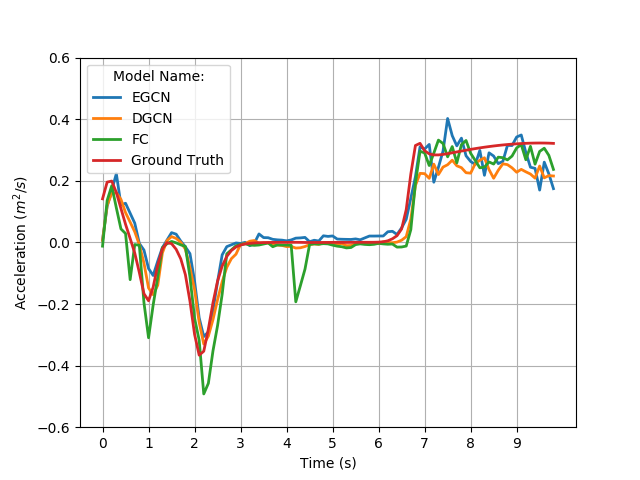}
\label{fig:subim2}
\caption{Fully-connected models}
\end{subfigure}
\begin{subfigure}{0.4\linewidth}
\centering
\includegraphics[width=1\linewidth, ]{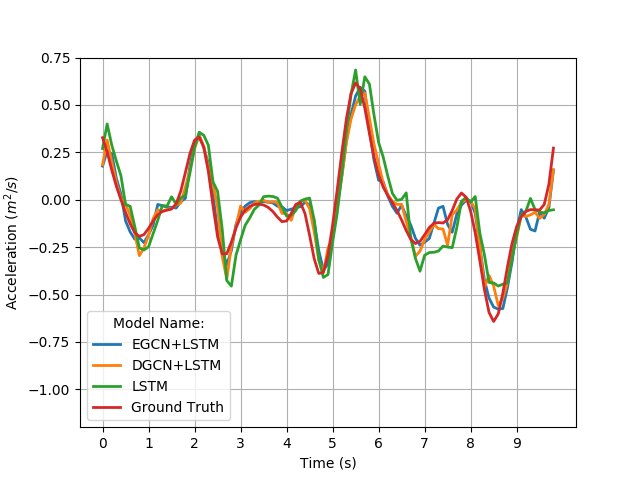}
\label{fig:subim3}
\caption{LSTM models}
\end{subfigure}
\begin{subfigure}{0.4\linewidth}
\centering
\includegraphics[width=1\linewidth, ]{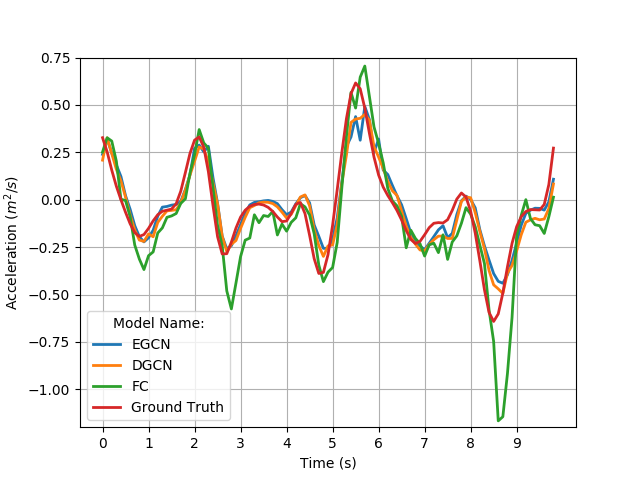}
\label{fig:subim4}
\caption{Fully-connected models}
\end{subfigure}
\caption{Simulated Trajectories For All Models (Orignial GCN and GAT models are excluded for their bad performance)}
\label{fig:sim_trajectory}
\end{figure*}
Figure \ref{fig:sim_trajectory} shows the sample simulated trajectories by models, including adapted GCN models and non-GCN models. It can be seen that non-GCN models predict poorly if the ground truth trajectory includes a long period of acceleration values that are very close to zero while GCN models is able to generate smooth trajectories close to the ground truth. In addition, non-GCN models are prone to predict extreme acceleration values, which is compensated by oscillation of acceleration values.

\section{Discussion}
Our experiments are designed to answer the following research questions:
%\begin{quote}
    \begin{itemize}
        \item Does GCN improve model performance and are our adaptations to GCN beneficial?
        \item Does including LSTM increase prediction quality?
        \item Why do GAT models fail to generate realistic trajectories?
    \end{itemize}
%end{quote}

First, we discover that we improve GCN's performance when we delete self-connections and apply different weights to the central nodes and their surrounding nodes. For GCN base model, we reduced velocity RMSE by $58.5\%$ at $10$ seconds horizon and negative headway occurrence by $17\%$ during simulation. For GCN with LSTM model, we reduced its $10$ seconds horizon velocity RMSE by $50.8\%$ and negative headway occurrence by $15\%$.

Second, our experiments demonstrated that GCNs improve model performance. GCN models are able to generate smooth and robust trajectories close to the ground truth. For both LSTM and fully-connected models, the non-GCN baseline model is outperformed by its GCN couterparts, in general. Note that, in the experiments, our GCN models and non-GCN models share the same number of hidden layers and the same number of neurons in each hidden layer. Compared with non-GCN fully-connected model, our EGCN model reduced the negative headway occurrence rate from $0.08$ to $0$, $10$ seconds horizon velocity RMSE by $59.6\%$. Compared with non-GCN LSTM, our DGCN with LSTM reduced the negative headway occurrence rate from $0.02$ to $0$, jerk sign inversions from $13.7$ to $7.3$ and $10$ seconds horizon velocity RMSE by $22.7\%$. This trend can also be observed in Figure \ref{fig:sim_trajectory}. The multi-layer GCN's ability to capture multitudes of interactions between vehicles hierarchically improves model's prediction quality in terms of our evaluation metrics.

In general, we find that adding LSTM structure improves model prediction quality. Among all models, the best model is DGCN with LSTM. During simulation, this model is able to generate robust and smooth driving trajectories with $0$ negative headway, $7.3$ jerk sign inversions and $0.256$ for 10-second horizon velocity RMSE.

GATs utilize self-attention to assign attentional weights $\alpha_{ij}$ for neighbouring node $j$. The attentional weights $\alpha_{ij}$ indicate the relationship between the central node and its surrounding nodes. Following \cite{bahdanau2014neural}, we investigated $\alpha_{ij}$ to understand why GAT models fail in our experiment. The investigation shows that the relational kernel in the baseline GAT models fails to learn the relationships between central nodes and their surrounding nodes. From the sample in the vehicle $829$ at the frame $2373$, the relational kernel in the second GAT layer of the GAT model assigns the same weights to every node: vehicle $829$ initially has two neighbours, $818$ and $835$. The attentional weights for each node, including the central node $829$, is $0.333$. Later in the trajectory, another vehicle $795$ approaches the ego-vehicle $829$ and the attentional weights assigned to all 4 nodes are $0.25$.

\section{Conclusion}
In this paper, we propose the use of graphs defined by the spatial relationships between vehicles, to model traffic. We further build GCN models, operating on graphs, to predict future acceleration distributions. We propose two GCN models adapted from the state-of-art GCN and studied the effectiveness of LSTM architectures in our prediction models. Our resulting frameworks outperform others on the task of acceleration prediction.

While our proposed methods have been shown to improve prediction performance, much work remains to be done. This work can be extended to prediction in two dimensions, which is an important problem in autonomous driving. At the same time, it will be interesting to evaluate different graph construction strategies, such as strategies that include multiple layers of relationships.

\bibliographystyle{IEEEtran}
\bibliography{mypaper}

\end{document}